# An Alternative Proof Method for Possibilistic Logic and its Application to Terminological Logics

Bernhard Hollunder
German Research Center for Artificial Intelligence (DFKI)
Stuhlsatzenhausweg 3, D-66123 Saarbrücken, Germany
e-mail: hollunder@dfki.uni-sb.de

## Abstract

Possibilistic logic, an extension of first-order logic, deals with uncertainty that can be estimated in terms of possibility and necessity measures. Syntactically, this means that a first-order formula is equipped with a possibility degree or a necessity degree that expresses to what extent the formula is possibly or necessarily true. Possibilistic resolution yields a calculus for possibilistic logic which respects the semantics developed for possibilistic logic.
A drawback, which possibilistic resolution inherits from classical resolution, is that it may not terminate if applied to formulas belonging to decidable fragments of first-order logic. Therefore we propose an alternative proof method for possibilistic logic. The main feature of this method is that it completely abstracts from a concrete calculus but uses as basic operation a test for classical entailment.
We then instantiate possibilistic logic with a terminological logic, which is a decidable subclass of first-order logic but nevertheless much more expressive than propositional logic. This yields an extension of terminological logics towards the representation of uncertain knowledge which is satisfactory from a semantic as well as algorithmic point of view.

## 1 Introduction

There have been many proposals for the treatment of uncertainty in Artificial Intelligence, in particular for expert systems and knowledge representation systems (for an overview see, e.g., [17, 14]). If uncertainty can be estimated in terms of possibility and necessity measures (as used in the framework of possibility theory [22]) possibilistic logic is a promising candidate. In fact, a basic feature of possibilistic logic is its ability to model states of knowledge ranging from complete information to total ignorance by expressing lower bounds for the possibility or necessity of some piece of knowledge. This allows, for instance, to distinguish between the total lack of certainty in the truth of a proposition and the certainty that the proposition is false.

From a syntactical point of view, possibilistic logic employs closed first-order formulas which are equipped with a possibility degree or a necessity degree: A weight $\Pi\alpha$ (resp. $N\alpha$) attached to a formula $p$ models to what extent $p$ is possibly (resp. necessarily) true, where $\alpha$ ranges between 0 and 1. To express, for example, that $p$ is likely to be true one may use the necessity-valued formula $(p, N0.7)$, whereas one may write $(p, \Pi0.9)$ to model that $p$ is to a high degree possible, but not certain at all.

Recently, a semantics for possibilistic logic has been presented for the general case where possibility- as well as necessity-valued formulas are allowed (cf. [15]). The semantics is based on fuzzy sets of interpretations, i.e., with each classical interpretation $\omega$ of the language associated with the first-order formulas occurring in a set of possibilistic formulas a value $\pi(\omega)$ between 0 and 1 is assigned. The possibility and necessity of a formula $p$ is then given by $\Pi(p) = \sup\{\pi(\omega) \mid \omega \models p\}$ and $N(p) = 1 - \Pi(\neg p)$. A fuzzy set of interpretations satisfies a possibilistic formula $(p, \Pi\alpha)$ (resp. $(p, N\alpha)$) iff $\Pi(p) \geq \alpha$ (resp. $N(p) \geq \alpha$). Entailment is then straightforwardly defined as follows: A possibilistic formula $\phi$ is a logical consequence of a possibilistic knowledge base $\Phi$, i.e., a set of possibilistic formulas, iff every fuzzy set of interpretations satisfying each element in $\Phi$ also satisfies $\phi$.

The entailment of a possibilistic formula from a possibilistic knowledge base can be checked mechanically on the basis of possibilistic resolution—an extension of the well-known resolution principle to possibilistic logic—which has been introduced in [8]. If applications of the possibilistic resolution rule to $\Phi \cup \{(\neg p, N1)\}$ yield a derivation of an empty possibilistic clause $(\square, v)$ then $\Phi$ entails $(p, v)$, where $v$ is either $\Pi\alpha$ or $N\alpha$ for some $\alpha \in [0, 1]$.

A drawback, which possibilistic resolution inherits from classical resolution, is that it may not terminate if applied to formulas belonging to decidable frag-ments of first-order logic. In fact, if the input formulas contain existential quantifiers in the scope of universal quantifiers, the (Skolem) function symbols that result from transforming these formulas into clause form may cause non-termination of standard resolution (and thus possibilistic resolution).[1] Moreover, the transformation of possibilistic formulas into clause form yields another problem: A set of possibility-valued formulas cannot always be transformed into an "equivalent" set of clauses—not even for the propositional case (cf. [15], Section 3.1).

For these reasons we propose an alternative proof method for possibilistic logic. The main feature of this method is that it completely abstracts from a concrete calculus (such as the resolution or tableaux calculus), and instead uses as basic operation a test for classical entailment. If this test is effective for a given fragment of first-order logic, we show that possibilistic reasoning is also decidable for this fragment. Additionally, if one has an algorithm that solves the entailment problem, our proof method automatically yields an algorithm realizing possibilistic entailment. We prove that the proposed method is sound and complete (for the general case where both possibility- and necessity-valued formulas are allowed) with respect to the semantics of possibilistic logic.

We then show how our method can be utilized to obtain decision procedures for a possibilistic extension of terminological knowledge representation formalisms, also called terminological logics. These formalisms, which are employed in terminological representation systems such as BACK [18], CLASSIC [5], KRIS [3], or LOOM [16] are in general decidable fragments of first-order logic, but are nevertheless expressive enough to define the relevant concepts of a problem domain. This is done by building complex concepts from primitive concepts (unary predicates) and roles (binary predicates) with the help of operations provided by the concept language of the particular formalism. For example, if we assume that *person* and *car* are concepts and that *owns* is a role, the concept *person*⊓∃*owns.car* describes the set of all persons having some car. Additionally, objects (or individuals) can be introduced by stating that an object is instance of a concept (e.g., *Tom:person*), or that two objects are related by a role (e.g., (*Tom, car_7*):*owns*).

Several approaches have already been proposed to enhance the expressivity of terminological formalisms with (some form of) uncertainty (e.g., probabilistic implications between concepts [10] or subsumption between fuzzy concepts [20]). The approach that comes nearest to ours, is described in [21]. It outlines an architecture for incorporating approximate reasoning into terminological systems. The main problem of this approach, however, is that its behavior is only described informally, i.e., neither a complete semantics nor algorithms for the main inference problems are given.

An extension of terminological formalisms towards the representation of uncertain knowledge which is satisfactory both from a semantic and algorithmic point of view can be obtained by instantiating possibilistic logic with a terminological logic. This means that we do not allow arbitrary first-order formulas in possibilistic formulas, but only those which can be formed by a particular terminological formalism. To be more precise, in the possibilistic extension we present one can, on the one hand, state plausible rules between concepts. For example, the rule

(*person* ⊓ *rich* → ∃*owns.Porsche*, Π0.7),

expresses that "rich persons are likely to own a Porsche." Of course, universally valid rules, i.e., strict implications between concepts such as "every Porsche is car" can be formulated by using the maximal necessity value N1. On the other hand, one can express uncertain knowledge concerning particular objects by adding possibility or necessity values to formulas expressing concept and role instanceships.

This approach has not only the advantage of being semantically sound. It also provides one with decision procedures for the basic inference problems (e.g., possibilistic entailment) which are sound and complete with respect to the semantics for possibilistic logic. These decision procedures can immediately be obtained by instantiating our proof method with inference algorithms for terminological logics as, for example, described in [6, 2].

The paper is organized as follows. In Section 2 we introduce syntax and semantics of possibilistic logic. The alternative proof method and the proof of its soundness and completeness are given in Section 3. Finally, in Section 4, we propose a possibilistic extension of terminological logics.

## 2 Possibilistic Logic

This section reviews possibilistic logic.[2] We start with introducing the syntax for possibilistic formulas, and then we recapitulate the semantics for possibilistic logic as defined in [15]. Finally, possibilistic resolution, a proof calculus for possibilistic logic, is presented.

A *possibilistic formula* is either a pair $(p, \Pi\alpha)$ or $(p, N\alpha)$ where $p$ is a closed first-order formula and $\alpha \in [0, 1]$ is a real number. A finite set of possibilistic formulas is called a *possibilistic knowledge base*.

Intuitively, a possibility-valued formula $(p, \Pi\alpha)$ (resp. necessity-valued formula $(p, N\alpha)$) expresses that $p$ is possibly (resp. necessarily) true at least to degree $\alpha$.

---

[1] It should be noted that the resolution calculus can be modified such that it yields decision procedures for various decidable fragments of first-order logic (see e.g. [19]). However, it is not yet clear whether or not this approach can be extended to the possibilistic case.

[2] For a more thorough introduction consult [7].

Let $\Omega$ be the set of interpretations of a first-order language. A *possibility distribution* $\pi$ on $\Omega$ is a mapping from $\Omega$ to $[0,1]$ such that $\pi(\omega) = 1$ for some $\omega \in \Omega$.

Note that the normalization requirement, i.e., $\pi(\omega) = 1$ for some $\omega \in \Omega$, guarantees that there is at least one world which could be considered the real one. Every possibility distribution $\pi$ on $\Omega$ induces two functions, denoted by $\Pi'$ and $N'$, mapping elements of

$$\widehat{\Phi} = \{p \mid (p, N\alpha) \in \Phi \text{ or } (p, \Pi\alpha) \in \Phi\}$$

to $[0,1]$. These functions, called possibility measure and necessity measure, are defined as follows.

Let $\pi$ be a possibility distribution on a set $\Omega$ of interpretations. The functions $\Pi'$ and $N'$ defined by

- $\Pi'(p) = \sup\{\pi(\omega) \mid \omega \in \Omega \text{ and } \omega \models p\}$
- $N'(p) = \inf\{1 - \pi(\omega) \mid \omega \in \Omega \text{ and } \omega \not\models p\}$

are called *possibility measure* and *necessity measure*, where $\sup\{\} := 0$ and $\inf\{\} := 1$.

A direct consequence of the definition is $\Pi'(p \vee q) = \max\{\Pi'(p), \Pi'(q)\}$ and $\Pi'(p \wedge q) \leq \min\{\Pi'(p), \Pi'(q)\}$, which in fact shows that the possibility measure is in accordance with the basic axioms of possibility theory (cf. [22]). Moreover, by duality of the measures $\Pi'$ and $N'$, i.e., $\Pi'(p) = 1 - N'(\neg p)$, we have for the necessity measure $N'(p \wedge q) = \min\{N'(p), N'(q)\}$ and $N'(p \vee q) \geq \max\{N'(p), N'(q)\}$. If a (first-order) formula $p \rightarrow q$ is valid, i.e. $\{p\} \models q$, it is easy to verify that $\Pi'(p) \leq \Pi'(q)$ and $N'(p) \leq N'(q)$; furthermore $\Pi'(\top) = N'(\top) = 1$ for any tautology $\top$, and $\Pi'(\bot) = N'(\bot) = 0$ for any inconsistent formula $\bot$.

In possibilistic logic, the notions of satisfaction and entailment are defined with respect to possibility distributions.

A possibility distribution $\pi$ on a set $\Omega$ of interpretations *satisfies* a possibilistic formula $(p, \Pi\alpha)$, written as $\pi \models (p, \Pi\alpha)$, iff $\Pi'(p) \geq \alpha$, and it *satisfies* $(p, N\alpha)$, written as $\pi \models (p, N\alpha)$, iff $N'(p) \geq \alpha$. A possibility distribution $\pi$ on $\Omega$ *satisfies* a possibilistic knowledge base $\Phi$ iff $\pi \models \phi$ for all $\phi \in \Phi$. Finally, we say that a possibilistic formula $\phi$ is *entailed* by a possibilistic knowledge base $\Phi$, denoted by $\Phi \models \phi$, iff $\pi \models \phi$ for all $\pi$ such that $\pi \models \Phi$ holds.

Let us consider an example. Consider the possibilistic knowledge base

$$\Phi = \{(p, N0.8), (p \rightarrow q, N0.4), (q \rightarrow r, \Pi0.7)\}.$$

Then $\Phi$ entails the formula $(r, \Pi 0.7)$. To see this, let $\pi$ be a possibility distribution satisfying $\Phi$. We first observe that $\pi \models (q, N 0.4)$. In fact, since $\{p, p \rightarrow q\} \models q$ we conclude $N'(q) \geq N'(p \wedge (p \rightarrow q)) = \min\{N'(p), N'(p \rightarrow q)\} \geq 0.4$. Duality gives us $\Pi'(\neg q) \leq 0.6$. Furthermore, note that $0.7 \leq \Pi'(q \rightarrow r) = \Pi'(\neg q \vee r) = \max\{\Pi'(\neg q), \Pi'(r)\}$. Since $\Pi'(\neg q) \leq 0.6$ we can conclude that $\Pi'(r) \geq 0.7$. This shows that $\pi$ satisfies $(r, \Pi 0.7)$.

There are possibilistic knowledge bases which are not satisfied by any possibilistic distribution. For example, if $\Phi$ contains both $(p, N 0.7)$ and $(\neg p, N 0.4)$, one gets $\Pi'(\neg p) = \sup\{\pi(\omega) \mid \omega \models \neg p\} \leq 0.3$ and $\Pi'(p) = \sup\{\pi(\omega) \mid \omega \models p\} \leq 0.6$ for every possibility distribution $\pi$. But this means that $\pi(\omega) \leq 0.6$ for every $\omega$, which shows that the normalization requirement, i.e., $\pi(\omega) = 1$ for some $\omega$, is not satisfied. Thus $\Phi$ cannot be satisfied by any possibilistic distribution. This, of course, means that *every* possibilistic formula is entailed by $\Phi$. However, the fact that we have more confidence in the truth of $p$ than in the truth of $\neg p$ is not taken into account by the semantics just described.

To achieve this behavior, i.e., to block the entailment of a possibilistic formula from an "inconsistent" knowledge base, we need the notion of an absurd interpretation. To define this notion, we (temporarily) view an interpretation $\omega$ as a function that maps a first-order formula $p$ to an element of the set $\{0, 1\}$ such that $\omega(p) = 1$ if $\omega \models p$ and $\omega(p) = 0$ if $\omega \not\models p$.

The function that maps each formula $p$ to 1 is called the *absurd interpretation* and is denoted by $\omega_\bot$, i.e., $\omega_\bot(p) = 1$ for all formulas $p$. By abuse of notation, in the following we simply write $\omega_\bot \models p$ iff $\omega_\bot(p) = 1$.

Let $\Omega$ be a set of (classical) interpretations and let $\omega_\bot$ be the absurd interpretation. A possibility distribution is now a mapping from $\Omega_\bot := \Omega \cup \{\omega_\bot\}$ to $[0, 1]$ such that $\pi(\omega) = 1$ for some $\omega \in \Omega_\bot$. The *possibility measure* $\Pi$ and *necessity measure* N induced by a possibility distribution $\pi$ on $\Omega_\bot$ is defined by

- $\Pi(p) = \sup\{\pi(\omega) \mid \omega \in \Omega_\bot \text{ and } \omega \models p\}$ and
- $N(p) = \inf\{1 - \pi(\omega) \mid \omega \in \Omega_\bot \text{ and } \omega \not\models p\}$.

Observe that $\Pi(p) = \max\{\Pi'(p), \pi(\omega_\bot)\}$ and $N(p) = N'(p)$, which means that the duality between $\Pi$ and N can be expressed by $\Pi(p) = \max\{1 - N(\neg p), \pi(\omega_\bot)\}$. Furthermore, it can easily be verified that the thus defined possibility and necessity measures satisfy the basic axioms of possibility theory.

Satisfaction and entailment are defined as before except that we now consider possibility distributions on $\Omega_\bot$ (instead of $\Omega$).

In [15] it has already been mentioned that both semantics coincide for possibilistic knowledge bases that are "consistent." To be more precise, suppose that there is a possibilistic distributions $\pi$ on $\Omega$ satisfying $\Phi$. Then $\phi$ is entailed by $\Phi$ according to the first semantics if and only if $\phi$ is entailed by $\Phi$ according to the modified, inconsistency tolerant, semantics. Of course, both semantics differ in the case where $\Phi$ is inconsistent. Recall that $\Phi$ given by $\{(p, N0.7), (\neg p, N0.4)\}$ entails every possibilistic formula according to the first semantics. However, according to the inconsistency tolerant semantics we have $\Phi \models (p, N0.7)$ and $\Phi \models (\neg p, N0.4)$, but $\Phi \not\models (p, N\alpha)$ for $\alpha > 0.7$ and $\Phi \not\models (\neg p, N\alpha')$ for $\alpha' > 0.4$. This shows that one can no longer derive any possibilistic formula from an inconsistent possibilistic knowledge base.

A possibilistic knowledge base that is inconsistent according to the first semantics is *more or less inconsistent* according to the inconsistency tolerant semantics. For example, $\{(p, N\alpha), (\neg p, N\alpha)\}$ should be considered more inconsistent than $\{(p, N\beta), (\neg p, N\beta)\}$ if $\alpha > \beta$. To measure the strength of inconsistency the following definition has been introduced in [15].

The *inconsistency degree* of a possibilistic knowledge base $\Phi$, *Incons*($\Phi$), is defined as follows:

- If there is a possibility distribution $\pi$ on $\Omega_\bot$ such that $\pi \models \Phi$ and $\pi(\omega) = 1$ for some $\omega \in \Omega$, then $\Phi$ is *possibly inconsistent* and *Incons*($\Phi$) = $\Pi\alpha$ where $\alpha = \inf\{\pi(\omega_\bot) \mid \pi \models \Phi\}$. If *Incons*($\Phi$) = $\Pi 0$ we say that $\Phi$ is *completely consistent*.

- If for all possibility distributions $\pi$ on $\Omega_\bot$, $\pi \models \Phi$ implies $\pi(\omega) < 1$ for every $\omega \in \Omega$, $\Phi$ is *necessarily inconsistent* and *Incons*($\Phi$) = $N\alpha$ where $\alpha = \inf\{1 - \pi(\omega) \mid \omega \in \Omega \text{ and } \pi \models \Phi\}$.

To illustrate this definition let us consider some examples. To determine the inconsistency degree of $\Phi_1 = \{(p, N\alpha), (\neg p, \Pi\beta)\}$ we construct an appropriate possibility distribution $\pi$ on $\Omega_\bot$ satisfying $\Phi_1$. If $\pi \models \Phi_1$ then $\pi(\omega) \leq 1 - \alpha$ for every interpretation $\omega$ with $\omega \not\models p$ (because $N(p) = 1 - \sup\{\pi(\omega) \mid \omega \in \Omega_\bot \text{ and } \omega \not\models p\} \geq \alpha$). First assume that $\alpha + \beta \leq 1$. We observe that the possibility distribution defined by

$$\pi(\omega) = \begin{cases} 1 & \text{if } \omega \not\models \neg p \\ \beta & \text{if } \omega \not\models p \\ 0 & \text{if } \omega = \omega_\bot \end{cases}$$

satisfies $\Phi_1$. In fact, $N(p) = 1 - \sup\{\pi(\omega) \mid \omega \in \Omega_\bot \text{ and } \omega \not\models p\} = 1 - \beta \geq \alpha$ and $\Pi(\neg p) = \sup\{\pi(\omega) \mid \omega \in \Omega_\bot \text{ and } \omega \models \neg p\} \geq \beta$, which shows that $\pi \models \Phi_1$. Thus $\Phi_1$ is *completely consistent* if $\alpha + \beta \leq 1$.

Now assume that $\alpha + \beta > 1$. Recall that $\pi(\omega) \leq 1 - \alpha$ for every $\omega$ with $\omega \not\models p$, which shows that $\sup\{\pi(\omega) \mid \omega \in \Omega_\bot \text{ and } \omega \not\models p\} < \beta$ (since $1 - \alpha < \beta$). But this means that $\pi(\omega_\bot) \geq \beta$ for all $\pi$ satisfying $\Phi_1$ because $\Pi(\neg p) = \max\{\Pi'(\neg p), \pi(\omega_\bot)\} \geq \beta$. Since the possibility distribution defined by

$$\pi(\omega) = \begin{cases} 1 & \text{if } \omega \not\models \neg p \\ 1 - \alpha & \text{if } \omega \not\models p \\ \beta & \text{if } \omega = \omega_\bot, \end{cases}$$

satisfies $\Phi_1$, we can conclude that $\Phi_1$ is *possibly inconsistent* and *Incons*($\Phi_1$) = $\Pi\beta$.

An example for a *necessarily inconsistent* possibilistic knowledge base is $\Phi_2 = \{(p, N\alpha), (\neg p, N\beta)\}$ where $\alpha > 0$ and $\beta > 0$. It can easily be checked that *Incons*($\Phi_2$) = $N \min\{\alpha, \beta\}$.

The total ordering on possibility and necessity measures is defined by $\Pi\alpha \geq \Pi\alpha'$ iff $\alpha \geq \alpha'$, $N\alpha \geq N\alpha'$ iff $\alpha \geq \alpha' > 0$, and $N\alpha \geq \Pi\alpha'$ iff $\alpha > 0$ and $\alpha' \leq 1$.

Thus we have that *Incons*($\Phi$) $\geq \Pi\alpha$ (resp. *Incons*($\Phi$) $\geq N\alpha$) implies *Incons*($\Phi$) $\geq \Pi\alpha'$ (resp. *Incons*($\Phi$) $\geq N\alpha'$) if $\alpha \geq \alpha'$. Furthermore, *Incons*($\Phi$) $\geq N\alpha$ implies *Incons*($\Phi$) $\geq \Pi\alpha'$ if $\alpha > 0$ and $\alpha' \leq 1$. This definition is justified by the fact that if $\Phi$ is necessarily inconsistent, then $\pi(\omega_\bot) = 1$ for all $\pi$ satisfying $\Phi$. But this means that $1 = \inf\{\pi(\omega_\bot) \mid \pi \models \Phi\} > \alpha$ for any inconsistency degree $\Pi\alpha$.

The following proposition, which has been proved in [15], shows that the entailment problem in possibilistic logic can be reduced to the problem of determining the inconsistency degree of a possibilistic knowledge base, and vice versa.

**Proposition 2.1 (Lang, Dubois, and Prade)** *Let $\Phi$ be a possibilistic knowledge base. Then:*

- $\Phi \models (p, \Pi\alpha)$ *iff* *Incons*($\Phi \cup \{(\neg p, N1)\}$) $\geq \Pi\alpha$,

- $\Phi \models (p, N\alpha)$ *iff* *Incons*($\Phi \cup \{(\neg p, N1)\}$) $\geq N\alpha$.

In order to determine (lower bounds for) the inconsistency degree of a possibilistic knowledge base the resolution principle has been extended such that it can be applied to possibilistic formulas (see, e.g., [15]). Let $(c, v)$, $(c', v')$ be possibilistic formulas, where $c, c'$ are first-order formulas in clause form and $v, v'$ are possibility or necessity degrees. The possibilistic resolution rule allows the derivation of a possibilistic formula $(res(c, c'), v \circ v')$, where $res(c, c')$ is a classical resolvent of $c, c'$, and $\circ$ is defined as $N\alpha \circ N\alpha' = N \min\{\alpha, \alpha'\}$, $\Pi\alpha \circ \Pi\alpha' = \Pi 0$, and

$$N\alpha \circ \Pi\alpha' = \Pi\alpha' \circ N\alpha = \begin{cases} \Pi\alpha' & \text{if } \alpha + \alpha' > 1 \\ \Pi 0 & \text{else} \end{cases}$$

We notice that if a derived formula has the possibility degree $\Pi 0$, the formula does not carry any additional information and can therefore be discarded. This means in particular that the resolution rule need not be applied to two possibility-valued clauses.

If applications of the rule yield a derivation of an empty possibilistic clause $(\Box, v)$ from a set $\Phi$ of possibilistic clauses, a lower bound for the inconsistency degree of $\Phi$ is given by $v$, i.e., *Incons*($\Phi$) $\geq v$ (cf. [15]).

In [15] it has been shown that possibilistic resolution is sound and complete in the following sense: Let $\Phi$ be a set of possibility- and necessity-valued propositional clauses, or a set of necessity-valued first-order clauses. Then *Incons*($\Phi$) $\geq v$ iff there is a derivation of an empty possibilistic clause $(\Box, v)$ from $\Phi$ by applications of the possibilistic resolution rule.

Although possibilistic resolution has this nice property, the overall calculus, i.e, transforming arbitrary possibilistic formulas into clause form and then applying the possibilistic resolution rule, has some drawbacks. In the presence of possibility-valued formulas it is in general not possible to transform a set of possibilistic formulas into a set of possibilistic clauses which have the *same* inconsistency degree (see [15], Section 3.1). Also standard resolution may not terminate even if applied to decidable fragments of first-order logic. This, of course, means that possibilistic resolution does in general not yield a decision procedure for a possibilistic extension of terminological logics.

## 3   An Alternative Proof Method for Possibilistic Logic

This section describes an alternative method for solving the entailment problem in possibilistic logic and for determining the inconsistency degree of a possibilistic knowledge base. The main feature of this method is that it completely abstracts from a concrete calculus, but uses as basic operation a test for classical entailment. If this test is effective for a given fragment of first-order logic, we will see that possibilistic reasoning is also decidable for this fragment.

In the following we assume that the possibility and necessity degree of a possibilistic formula is not equal to zero. This assumption is justified by the fact that by definition $\Pi(p) \geq 0$ and $N(p) \geq 0$ hold, which shows that every possibility distribution satisfies formulas of the form $(p, \Pi 0)$ or $(p, N0)$. Hence such formulas do not carry any additional information and can therefore be discarded from possibilistic knowledge bases.

Let $\Phi$ be a possibilistic knowledge base and let $\alpha \in [0,1]$. We denote by $\Phi_\alpha$ (resp. $\Phi^\alpha$) the first-order formulas of necessity-valued formulas in $\Phi$ that have a value greater (resp. strictly greater) than $\alpha$, i.e.,

- $\Phi_\alpha := \{p \mid (p, N\alpha') \in \Phi,\ \alpha' \geq \alpha\}$ and
- $\Phi^\alpha := \{p \mid (p, N\alpha') \in \Phi,\ \alpha' > \alpha\}$.

These abbreviations are quite useful to give an alternative characterization of possibilistic entailment. Let $\Phi$ be a possibilistic knowledge base, let $p$ be a first-order formula, and let $0 < \alpha \leq 1$. We show that

- $\Phi \models (p, N\alpha)$ iff $\Phi_\alpha \models p$
- $\Phi \models (p, \Pi\alpha)$ iff
  - $\Phi^0 \models p$ or
  - there is some $(q, \Pi\beta) \in \Phi$ such that $\beta \geq \alpha$ and $\Phi^{1-\beta} \cup \{q\} \models p$.

This means that $(p, N\alpha)$ is entailed by $\Phi$ iff the first-order formulas of necessity-valued formulas in $\Phi$ whose value is not less than $\alpha$ classically entail $p$. For possibility-valued formulas the situation is slightly more complex: $(p, \Pi\alpha)$ is a possibilistic consequence of $\Phi$ iff (1) the first-order formulas of necessity-valued formulas in $\Phi$ classically entail $p$, or (2) there is a possibility-valued formula $(q, \Pi\beta)$, $\beta \geq \alpha$, in $\Phi$ such that $q$ together with the first-order formulas of necessity-valued formulas in $\Phi$ whose value is strictly greater than $1-\beta$ yield a classical proof for $p$.

Due to lack of space we omit the soundness and completeness proof for *necessity-valued* formulas (see [11]).

**Lemma 3.1** *Let $\Phi$ be a possibilistic knowledge base and let $(p, \Pi\alpha)$ be a possibilistic formula with $\alpha > 0$. If $\Phi^0 \models p$ or there is some $(q, \Pi\beta) \in \Phi$ such that $\beta \geq \alpha$ and $\Phi^{1-\beta} \cup \{q\} \models p$, then $\Phi \models (p, \Pi\alpha)$.*

*Proof.* Assume that $\Phi^0 \models p$ holds. There is a subset $\{(p_1, N\alpha_1), \ldots, (p_n, N\alpha_n)\}$ of $\Phi$ such that $\{p_1, \ldots, p_n\} \models p$ and $\min\{\alpha_1, \ldots, \alpha_n\} > 0$. This shows that $N(p) > 0$. Thus, if $\pi$ is a possibility distribution satisfying $\Phi$, we can conclude that for all $\omega \in \Omega_\perp$, $\omega \not\models p$ implies $\pi(\omega) < 1$. Because of the normalization requirement there is an interpretation $\omega'$ such that $\pi(\omega') = 1$. Since $\omega' \models p$ it follows that $\Pi(p) = 1$, and hence $\pi \models (p, \Pi\alpha)$. Thus we can conclude that $\Phi \models (p, \Pi\alpha)$.

Now assume that there is some $(q, \Pi\beta) \in \Phi$ such that $\beta \geq \alpha$ and $\Phi^{1-\beta} \cup \{q\} \models p$. Thus there is a subset $\{(p_1, N\alpha_1), \ldots, (p_n, N\alpha_n)\}$ of $\Phi$ such that $\{p_1, \ldots, p_n, q\} \models p$ and $\alpha_i > 1 - \beta$ for all $i$, $1 \leq i \leq n$. Let $\pi$ be a possibility distribution on $\Omega_\perp$ such that $\pi \models \Phi$. We show that $\pi$ satisfies $(p, \Pi\alpha)$. Let us recall that $\Pi(q) = \max\{\Pi'(q),\ \pi(\omega_\perp)\} \geq \beta \geq \alpha$.
*Case 1:* $\Pi(q) = \pi(\omega_\perp)$. Then $\Pi(p) = \sup\{\pi(\omega) \mid \omega \in \Omega_\perp \text{ and } \omega \models p\} \geq \pi(\omega_\perp) \geq \alpha$, which shows that $\pi$ satisfies $(p, \Pi\alpha)$.
*Case 2:* $\Pi(q) \neq \pi(\omega_\perp)$. Hence $\Pi(q) = \Pi'(q)$. First we show that $\Pi'(q \wedge p_1 \wedge \ldots \wedge p_n) \geq \beta$. Observe that

$\beta \leq \Pi'(q)$
$= \Pi'((q \wedge p_1 \wedge \ldots \wedge p_n) \vee (q \wedge \neg(p_1 \wedge \ldots \wedge p_n)))$
$= \Pi'((q \wedge p_1 \wedge \ldots \wedge p_n) \vee (q \wedge \neg p_1) \vee \ldots \vee (q \wedge \neg p_n))$
$= \max\{\Pi'(q \wedge p_1 \wedge \ldots \wedge p_n),$
$\quad\quad \Pi'(q \wedge \neg p_1), \ldots, \Pi'(q \wedge \neg p_n)\},$

and thus it remains to be shown that $\Pi'(q \wedge \neg p_i) < \beta$ for all $i$, $1 \leq i \leq n$. Since $N(p_i) = N'(p_i) \geq \alpha_i$ (which follows from the fact that $\pi$ satisfies every $(p_i, N\alpha_i)$) we have $\Pi'(\neg p_i) \leq 1 - \alpha_i$. Recall that $\alpha_i > 1 - \beta$, which shows that $\Pi'(\neg p_i) < \beta$, and therefore $\Pi'(q \wedge \neg p_i) < \beta$ for $i$, $1 \leq i \leq n$. Thus we can conclude that $\Pi'(q) = \Pi'(q \wedge p_1 \wedge \ldots \wedge p_n) \geq \beta$. Since $\Pi(q \wedge p_1 \wedge \ldots \wedge p_n) \geq \Pi'(q \wedge p_1 \wedge \ldots \wedge p_n)$ and $\{p_1, \ldots, p_n, q\} \models p$ we know that $\Pi(p) \geq \beta \geq \alpha$. Thus $\pi$ satisfies $(p, \Pi\alpha)$. □

Before we prove completeness we need one more definition and a proposition. Let $\Phi$ be a possibilistic knowledge base only containing necessity-valued formulas. The *canonical possibility distribution*[3] $\pi$ on $\Omega_\perp$ for $\Phi$ is defined by $\pi(\omega) = 1 - \max\{\alpha \mid (p, N\alpha) \in \Phi \text{ and } \omega \not\models p\}$, where $\max\{\} := 0$.

Notice that $\pi(\omega_\perp) = 1 - \max\{\alpha \mid (p, N\alpha) \in \Phi \text{ and } \omega_\perp \not\models p\} = 1$, which shows that the canonical possibility distribution satisfies the normalization constraint.

**Proposition 3.2** *Let $\Phi$ be a finite set of necessity-valued formulas and let $\pi$ be the canonical possibility distribution for $\Phi$. Then:*

1. $\pi(\omega) \leq 1 - \alpha$ *if* $(p, N\alpha) \in \Phi$ *and* $\omega \not\models p$.

2. $\pi$ *satisfies* $\Phi$.

The proof is an immediate consequence of the definition (cf. [11]).

---

[3]Such a distribution is also called *least specific possibility distribution* in [4].

**Lemma 3.3** *Let $\Phi$ be a possibilistic knowledge base and let $(p, \Pi\alpha)$ be a possibilistic formula with $\alpha > 0$. If $\Phi \models (p, \Pi\alpha)$ then $\Phi^0 \models p$ or there is some $(q, \Pi\beta) \in \Phi$ such that $\beta \geq \alpha$ and $\Phi^{1-\beta} \cup \{q\} \models p$.*

**Proof.** Assume that $\Phi \models (p, \Pi\alpha)$ for some $\alpha > 0$. If $\Phi^0 \models p$ we are done. Thus assume that $\Phi^0 \not\models p$. We show that there is a formula $(q, \Pi\beta)$ in $\Phi$ such that $\beta \geq \alpha$ and $\Phi^{1-\beta} \cup \{q\} \models p$.

Suppose to the contrary that for all $(q, \Pi\beta)$ in $\Phi$ such that $\beta \geq \alpha$ we have $\Phi^{1-\beta} \cup \{q\} \not\models p$. In the following we construct a possibility distribution $\pi'$ such that $\pi'$ satisfies $\Phi \cup \{(\neg p, \mathrm{N}1)\}$ and $\pi'(\omega_\bot) < \alpha$. But this means that $Incons(\Phi \cup \{(\neg p, \mathrm{N}1)\}) < \Pi\alpha$, which shows that $\Phi \not\models (p, \Pi\alpha)$ (Proposition 2.1), thus contradicting the assumption that $\Phi \models (p, \Pi\alpha)$ holds.

Let $\pi$ be the canonical possibility distribution for $\{(p', \mathrm{N}\alpha') \mid (p', \mathrm{N}\alpha') \in \Phi\} \cup \{(\neg p, \mathrm{N}1)\}$. The possibility distribution $\pi'$ for $\Phi \cup \{(\neg p, \mathrm{N}1)\}$ is constructed as follows:

$$\pi'(\omega) = \begin{cases} \pi(\omega) & \text{if } \omega \not\models \Phi^0 \cup \{\neg p\} \\ 1/2(\alpha + \gamma) & \text{if } \omega = \omega_\bot \\ 1 & \text{otherwise,} \end{cases}$$

where $\gamma = \max\{\beta \mid (r, \mathrm{N}\beta) \in \Phi \text{ and } \beta < \alpha\}$.

We first show that the normalization constraint is satisfied and that $\pi'(\omega_\bot) < \alpha$. On the one hand, we assumed that $\Phi^0 \not\models p$, which means that there is some interpretation $\omega'$ such that $\omega' \models \Phi^0 \cup \{\neg p\}$. Hence we have $\pi'(\omega') = 1$. On the other hand, we observe that $\gamma < \alpha$, which means that $\pi'(\omega_\bot) < \alpha$.

Next we prove that $\pi'$ satisfies $\Phi \cup \{(\neg p, \mathrm{N}1)\}$, i.e., we show that $\pi' \models \phi$ for every $\phi \in \Phi \cup \{(\neg p, \mathrm{N}1)\}$.

(1.) $(p', \mathrm{N}\alpha') \in \Phi \cup \{(\neg p, \mathrm{N}1)\}$:
Then:

$$\begin{aligned} N(p') &= 1 - \sup\{\pi'(\omega) \mid \omega \in \Omega_\bot, \omega \not\models p'\} \\ &= 1 - \sup\{\pi(\omega) \mid \omega \in \Omega_\bot, \omega \not\models p'\} \\ &\geq \alpha', \end{aligned}$$

which shows that $\pi' \models (p', \mathrm{N}\alpha')$.

(2.) $(p', \Pi\alpha') \in \Phi$ where $\alpha' \geq \alpha$.
Note that $\Pi(p') = \sup\{\pi'(\omega) \mid \omega \in \Omega_\bot, \omega \models p'\}$, and thus it suffices to show that there is some $\omega' \in \Omega_\bot$ such that $\omega' \models p'$ and $\pi'(\omega') \geq \alpha'$.

*Case 1:* There is an interpretation $\omega'$ different from $\omega_\bot$ such that $\omega' \models \Phi^0 \cup \{p', \neg p\}$. Then $\pi'(\omega') = 1$ (definition of $\pi'$), which shows that $\Pi(p') \geq \pi'(\omega') = 1 \geq \alpha'$. Thus $\pi' \models (p', \Pi\alpha')$.

*Case 2:* Now suppose that $\omega \not\models \Phi^0 \cup \{p', \neg p\}$ for every interpretation $\omega$ different from $\omega_\bot$. Recall that we assumed that $\Phi^{1-\alpha'} \cup \{p'\} \not\models p$. This means that $\Phi^{1-\alpha'} \cup \{p', \neg p\}$ is consistent, and hence there is some interpretation $\omega' \neq \omega_\bot$ such that $\omega' \models \Phi^{1-\alpha'} \cup \{p', \neg p\}$. Since we assumed that $\omega \not\models \Phi^0 \cup \{p', \neg p\}$ for every interpretation $\omega$ different from $\omega_\bot$, we can conclude that there is some $(p'', \mathrm{N}\alpha'') \in \Phi$ such that $\omega' \not\models p''$ and $\alpha'' \leq 1 - \alpha'$.

Since $\pi'(\omega') = \pi(\omega')$ (definition of $\pi'$), it remains to be shown that $\pi(\omega') \geq \alpha'$. In fact,

$$\begin{aligned} \pi(\omega') &= 1 - \max\{\beta \mid (r, \mathrm{N}\beta) \in \Phi \cup \{(\neg p, \mathrm{N}1)\}, \omega' \not\models r\} \\ &= 1 - \max\{\beta \mid (r, \mathrm{N}\beta) \in \Phi \cup \{(\neg p, \mathrm{N}1)\}, \\ &\qquad\qquad \beta \leq 1 - \alpha', \omega' \not\models r\} \\ &\geq \alpha' \qquad (\text{since } \beta \leq 1 - \alpha'). \end{aligned}$$

Thus we have shown that $\Pi(p') \geq \pi'(\omega') = \pi(\omega') \geq \alpha'$ and therefore we can conclude that $\pi' \models (p', \Pi\alpha')$.

(3.) $(p', \Pi\alpha') \in \Phi$ where $\alpha' < \alpha$.
Since $\Pi(p') \geq \pi'(\omega_\bot)$ it suffices to show that $\pi'(\omega_\bot) \geq \alpha'$. In fact, $\pi'(\omega_\bot) = 1/2(\alpha + \gamma) \geq \alpha'$ (because $\gamma \geq \alpha'$ as well as $\alpha > \alpha'$).

Thus we have shown that $\pi'$ satisfies $\Phi \cup \{(\neg p, \mathrm{N}1)\}$, which concludes the proof. □

The previous two lemmata together with Lemma 3.1 and Lemma 3.4 of [11] establish the main result of this section.

**Theorem 3.4** *Let $\Phi$ be a possibilistic knowledge base, let $p$ be a first-order formula, and let $\alpha > 0$. Then*

- *$\Phi \models (p, \mathrm{N}\alpha)$ iff $\Phi_\alpha \models p$ and*
- *$\Phi \models (p, \Pi\alpha)$ iff*
  - *$\Phi^0 \models p$ or*
  - *there is some $(q, \Pi\beta) \in \Phi$ such that $\beta \geq \alpha$ and $\Phi^{1-\beta} \cup \{q\} \models p$.*

**Corollary 3.5** *Possibilistic entailment is decidable in those languages in which classical entailment is decidable.*

In the rest of this section we consider the problem of how to determine (with the help of Theorem 3.4) the inconsistency degree of a possibilistic knowledge base $\Phi$. By Proposition 2.1 we know that $\Phi \models (\bot, v)$ iff $Incons(\Phi \cup \{(\neg\bot, \mathrm{N}1)\}) \geq v$, and hence $\Phi \models (\bot, v)$ iff $Incons(\Phi) \geq v$, where $\bot$ is an inconsistent formula and $v$ is a necessity or possibility measure. Thus the problem is to find the maximal value $v$ such that $\Phi \models (\bot, v)$.

Let $\gamma := \min\{\alpha \mid (p, \mathrm{N}\alpha) \in \Phi\}$. First assume that $\Phi_\gamma \models \bot$. This means that $\Phi$ is necessarily inconsistent at least to degree $\gamma$. Observe that $\Phi_\alpha \supseteq \Phi_{\alpha'}$ iff $\alpha \leq \alpha'$. Hence, in order to determine the number $\alpha \in \{\beta \mid (p, \mathrm{N}\beta) \in \Phi\}$ such that $\Phi_\alpha \models \bot$ but $\Phi^\alpha \not\models \bot$ one can for instance apply a binary search algorithm (rather than testing for each element $\alpha \in \{\alpha \mid (p, \mathrm{N}\alpha) \in \Phi\}$ whether or not $\Phi_\alpha$ is inconsistent). The inconsistency degree of $\Phi$ is then given by $\mathrm{N}\alpha$.

Now assume that $\Phi_\gamma \not\models \bot$. If $\Phi^{1-\beta} \cup \{q\}$ is consistent for every $(q, \Pi\beta)$ in $\Phi$, we can conclude that $\Phi$ is completely consistent (which means that $Incons(\Phi) = \Pi 0$). Otherwise, the maximal number $\beta$ such that $(q, \Pi\beta) \in \Phi$ and $\Phi^{1-\beta} \cup \{q\}$ is inconsistent yields the inconsistency degree of $\Phi$, i.e., $Incons(\Phi) = \Pi\beta$. Note that if $(q, \Pi\beta)$ and $(q', \Pi\beta')$ are in $\Phi$ where $\beta \leq \beta'$, in general neither $Th(\{q\} \cup \Phi^{1-\beta}) \subseteq Th(\{q'\} \cup \Phi^{1-\beta'})$ nor $Th(\{q\} \cup \Phi^{1-\beta}) \supseteq Th(\{q'\} \cup \Phi^{1-\beta'})$ hold. This, however, means that one cannot employ binary search to determine the required value $\beta$.

To sum up, assume that $\Phi$ is a possibilistic knowledge base with $n$ formulas and $p$ is a first-order formula. Then one can determine the maximal number $\alpha$ such that $\Phi \models (p, N\alpha)$ holds with $\mathcal{O}(\log n)$ classical entailment tests. In contrast to this, one can determine with $\mathcal{O}(n)$ entailment tests the maximal number $\alpha$ such that $\Phi \models (p, \Pi\alpha)$ holds.

## 4 A Possibilistic Extension of Terminological Logics

This section describes an extension of terminological knowledge representation formalisms that handles uncertain knowledge and allows for approximate reasoning. This approach is not only satisfactory from a semantical point of view; it also provides sound and complete decision procedures for the basic inference problems. These algorithms can immediately be obtained by instantiating our proof method with the well-known inference algorithms for terminological logics.

### 4.1 Terminological knowledge representation

In the following we briefly introduce a particular terminological formalism, called $\mathcal{ALCN}$ (cf. [13]). Such a formalism can be used to define the relevant concepts of a problem domain. Relationships between concepts, for instance inclusion or disjointness axioms, can be expressed in the terminological part. The assertional part allows one to describe objects of the problem domain with respect to their relation to concepts and their interrelation with each other.

We assume two disjoint alphabets of symbols, called *primitive concepts* and *roles*. The set of *concepts* is inductively defined as follows. Every primitive concept is a concept. Now let $C$, $D$ be concepts already defined and let $R$ be a role. Then $C \sqcap D$ (conjunction), $C \sqcup D$ (disjunction), $\neg C$ (negation), $\forall R.C$ (value-restriction), $\exists R.C$ (exists-restriction), and $(\geq n\ R)$ and $(\leq n\ R)$ (number-restrictions) are concepts of the language $\mathcal{ALCN}$.

Concepts are usually interpreted as subsets of a domain and roles as binary relations over a domain. This means that primitive concepts (resp. roles) are considered as symbols for unary (resp. binary) predicates, and that concepts correspond to formulas with one free variable. Thus primitive concepts $A$ and roles $R$ are translated into atomic formulas $A(x)$ and $R(x,y)$,

where $x,y$ are free variables. The semantics of the concept-forming constructs is given by
$(C \sqcap D)(x) := C(x) \wedge D(x)$,
$(C \sqcup D)(x) := C(x) \vee D(x)$,
$(\neg C)(x) := \neg C(x)$,
$(\forall R.C)(x) := \forall y\ (R(x,y) \rightarrow C(y))$,
$(\exists R.C)(x) := \exists y\ (R(x,y) \wedge C(y))$,
$(\geq n\ R)(x) := \exists y_1, \ldots, y_n\ y_1 \neq y_2 \wedge y_1 \neq y_3 \wedge \ldots \wedge y_{n-1} \neq y_n \wedge R(x,y_1) \wedge \ldots \wedge R(x,y_n)$,
$(\leq n\ R)(x) := \forall y_1, \ldots, y_{n+1}\ R(x,y_1) \wedge \ldots \wedge R(x,y_{n+1}) \rightarrow y_1 = y_2 \vee y_1 = y_3 \vee \ldots \vee y_{n-1} = y_n$.

It should be noted that the formulas thus obtained belong to a restricted subclass of all first-order formulas with one free variable.

A terminological knowledge base is described by a set of inclusion axioms and—to introduce objects with respect to their relation to concepts and their interrelation with each others—a set of membership assertions.

To be more formal, let $C, D$ be concepts, $R$ be a role, and let $a, b$ be names for individuals, so-called *objects*. A *terminological axiom* is of the form $C \rightarrow D$, and expresses that every instance of $C$ is also an instance of $D$. To state that an object $a$ belongs to a concept $C$, or that two objects $a, b$ are related by a role $R$ one can use *assertions* having the form $C(a)$ or $R(a,b)$.

The semantics of a terminological axiom $C \rightarrow D$ is given by the formula $\forall x\ C(x) \rightarrow D(x)$, where $C(x), D(x)$ are the first-order formulas corresponding to the concepts $C, D$. To define the semantics of assertions we consider individual names as symbols for constants. In terminological systems one usually has a *unique name assumption*, which can be expressed by the formulas $a \neq b$ for all distinct individual names $a, b$. The formula corresponding to the assertion $C(a)$ (resp. $R(a,b)$) is obtained by replacing the free variable(s) in the formula corresponding to $C$ (resp. $R$) by $a$ (resp. $a, b$).

A *terminological knowledge base* is a pair $(\mathcal{T}, \mathcal{A})$ where $\mathcal{T}$ is a finite set of terminological axioms (the so-called *TBox*) and $\mathcal{A}$ is a finite set of assertions (the so-called *ABox*). Observe that a terminological knowledge base $(\mathcal{T}, \mathcal{A})$ can be viewed as a finite set of first-order formulas that can be obtained by taking the translations of the TBox and ABox facts, and the formulas expressing unique name assumption.

The basic inference services for terminological knowledge bases are defined as follows:

*Consistency checking:* Does there exist a model for a given terminological knowledge base $(\mathcal{T}, \mathcal{A})$ ?
*Subsumption problem:* Is a terminological axiom $C \rightarrow D$ entailed by $(\mathcal{T}, \mathcal{A})$, i.e., $(\mathcal{T}, \mathcal{A}) \models \forall x\ C(x) \rightarrow D(x)$ ?
*Instantiation problem:* Is an assertion $C(a)$ (resp. $R(a,b)$) entailed by $(\mathcal{T}, \mathcal{A})$, i.e., $(\mathcal{T}, \mathcal{A}) \models C(a)$ (resp. $(\mathcal{T}, \mathcal{A}) \models R(a,b)$) ?

It should be noted that these inference problems are decidable for most terminological logics.

### 4.2  The possibilistic extension

The possibilistic extension of the terminological formalism introduced in the previous subsection is obtained as follows: Each terminological axiom (resp. assertion) is equipped with a possibility or a necessity value and will be called *possibilistic terminological axiom* (resp. *possibilistic assertion*). A *possibilistic knowledge base* is now a set of possibilistic terminological axioms together with a set of possibilistic assertions.

In order to give some impression on the expressivity of the extended terminological language, let us consider two examples. The first one, which is taken from [21], is concerned with strict terminological axioms but uncertain assertions. Assume that $\mathcal{T}$ is given by

(father $\leftrightarrow$ man $\sqcap$ ($\geq$ 1 child), N1)
(successful_father $\leftrightarrow$ father $\sqcap$ $\forall$child.college_grad., N1),

where $(C \leftrightarrow D, \text{N1})$ is an abbreviation for the axioms $(C \rightarrow D, \text{N1})$ and $(D \rightarrow C, \text{N1})$. The first axiom expresses that someone is a father iff he is a man and has some child; the latter one states that someone is a successful father iff he is a father and all his children are college graduates.

First consider the (certain) assertions

$\mathcal{A} = \{$ (John: man $\sqcap$ ($\leq$ 2 child), N1),
((John, Philip): child, N1),
(Philip: college_grad., N1),
((John, Angela): child, N1),
(Angela: college_grad., N1) $\}$,

which state that John is a man having at most two children, that Philip and Angela are children of John, and that both are college graduates. Since Philip and Angela are the only children of John (because he has at most two children) and both children are college graduates, we can conclude that John is a successful father, i.e., the possibilistic assertion (John : successful_father, N1) is entailed by $(\mathcal{T}, \mathcal{A})$.

Now assume that it is only likely that Philip is a college graduate, which can be encoded by (Philip : college_grad., N0.8). Again, by possibilistic entailment we conclude that John is a successful father but, of course, only with a necessity degree of 0.8.

In the second example, possibility and necessity degrees are utilized to express *plausible rules*. Assume that the TBox $\mathcal{T}$ contains the following possibilistic axioms:

($\exists$owns.porsche $\rightarrow$ rich_person $\sqcup$ car_fanatic, N0.8)
(rich_person $\rightarrow$ golfer, $\Pi$0.7).

The first axiom expresses that it is rather certain that someone is either rich or a car fanatic if (s)he owns a Porsche. The second one states that rich persons are possibly golfers.

The assertional knowledge is given by the facts that Tom owns a Porsche 911 and that he is probably not a car fanatic, i.e.,

$\mathcal{A} = \{$ ((Tom, 911): owns, N1),
(911: porsche, N1),
(Tom: $\neg$car_fanatic, N0.7) $\}$.

We are interested in the question of whether or not Tom is a golfer. To answer the query observe that

$\{(\text{Tom}, 911) : \text{Owns}, 911 : \text{porsche}\} \models \text{Tom} : \exists \text{owns.porsche}$,

which shows that (Tom : $\exists$owns.porsche, N1) is entailed by $(\mathcal{T}, \mathcal{A})$. Hence, it can easily be verified that $(\mathcal{T}, \mathcal{A})^{1-0.7} \cup \{\text{Tom} : \text{rich\_person}\} \models \text{Tom} : \text{golfer}$. This shows that (Tom : golfer, $\Pi$0.7) is a possibilistic consequence of $(\mathcal{T}, \mathcal{A})$, which means that we have some reasons to believe that Tom is a golfer.

The following proposition shows that possibilistic reasoning restricted to the introduced terminological formalism is decidable. This result is an immediate consequence of Theorem 3.4 and the fact that the instantiation problem in $\mathcal{ALCN}$-knowledge bases is decidable (cf. [6, 1]).

**Proposition 4.1** *Let $\mathcal{T}$ be a finite set of possibilistic axioms and let $\mathcal{A}$ be a finite set of possibilistic assertions. It is decidable whether or not a possibilistic axiom (resp. possibilistic assertion) is entailed by $(\mathcal{T}, \mathcal{A})$.*

Almost all terminological systems do not allow arbitrary TBoxes, but only those that satisfy certain conditions (for instance, the left hand side of an axiom must be a primitive concept, and a primitive concept may appear at most once at the left hand side of an axiom). In [12, Chapter 7.3] it has been shown how to obtain more efficient inference procedures if possibilistic TBoxes satisfy the additional restrictions.

## 5  Conclusion

We have developed an alternative proof method for possibilistic logic which exploits the fact that possibilistic reasoning can be reduced to reasoning in classical, i.e. first-order, logic. Consequently, possibilistic reasoning is decidable for a fragment of first-order logic iff classical entailment is decidable for it. Moreover, if one has an algorithm solving the entailment problem, our method automatically yields an algorithm realizing possibilistic entailment which is sound and complete with respect to the semantics for possibilistic logic.

Furthermore, we have instantiated possibilistic logic with a terminological logic, which is a decidable fragment of first-order logic, but nevertheless much more expressive than propositional logic. This leads to an extension of terminological logics towards the representation of uncertain knowledge which is—in contrast to other approaches—satisfactory from a semantic point of view. Moreover, a sound and complete algorithm for possibilistic entailment in such an extension can be obtained by using inference procedures which have already been developed for terminological logics.

An interesting point for further research is to employ possibilistic logic in order to represent and reason with defaults in terminological formalisms. In fact, in [9, 4] it has been argued that possibilistic logic yields a good basis for nonmonotonic reasoning. Roughly speaking, the idea is as follows: If the necessity of a formula $p$ is greater than the necessity of $\neg p$ with respect to a set $\Phi$ of necessity-valued formulas, then infer nonmonotonically $p$ from $\Phi$. This intuitive definition in fact characterizes an appropriate nonmonotonic consequence relation as (1) the operator can be described in terms of preferential models, and (2) most of the axioms which a nonmonotonic operator should satisfy are met (see [9] on these points). The approach presented in [4], however, uses propositional logic and cannot directly be applied to the terminological case. One reason for this is that terminological default rules usually allow one to state that "$C$'s are normally $D$'s" where $C, D$ are concepts, i.e., first-order formulas with one free variable.

**Acknowledgements**
I would like to thank Werner Nutt who helped me to simplify notations, and Franz Baader, Detlef Fehrer, and Jörg Siekmann for helpful comments on a draft of this paper. This work has been supported by the German Ministry for Research and Technology (BMFT) under research contract ITW 92 01.